\documentclass[twoside]{article}

\usepackage[accepted]{aistats2020}
\usepackage{latexsym,amssymb,amsmath,ifthen,afterpage, amsfonts, mathtools}

\usepackage{enumerate}
\usepackage[utf8]{inputenc} 
\usepackage[T1]{fontenc}    
\usepackage{lmodern}
\usepackage{url}            
\usepackage{booktabs}       
\usepackage{nicefrac}       
\usepackage{microtype}      
\usepackage{algorithm}
\usepackage{algorithmic}
\usepackage{balance}

\usepackage{tikz} 
\usepackage{pgfplots}
\usepackage{pgfplotstable}
\usetikzlibrary{patterns}
\pgfplotsset{compat = newest}
\pgfkeys{/pgf/number format/set thousands separator = }
\definecolor{chocolate1}{RGB}{1,100,5}
\definecolor{chocolate2}{RGB}{145,65,12}

\newcommand{\Fig}[1]{Fig.~\ref{#1}}

\newcommand{\R}{\mathbb{R}}

\newcommand{\ZZ}{\mathbb{Z}}

\newcommand{\calA}{\mathcal{A}}
\newcommand{\calX}{\mathcal{X}}

\newcommand{\calU}{\mathcal{U}}

\newcommand{\calN}{\mathcal{N}}
\newcommand{\calW}{\mathcal{W}}
\newcommand{\EE}{\mathcal{E}}
\newcommand{\X}{{\bf X}}
\newcommand{\x}{{\bf x}}
\newcommand{\Y}{{\bf Y}}
\newcommand{\y}{{\bf y}}

\newcommand{\E}{\operatorname{E}}

\newcommand{\e}{{e'}}

\newcommand{\G}{\mathcal{G}}
\newcommand{\VV}{\mathcal{V}}
\newcommand{\HH}{\mathcal{H}}

\newcommand{\overbar}[1]{\mkern 1.5mu\overline{\mkern-1.5mu#1\mkern-1.5mu}\mkern 1.5mu}

\newcommand{\I}{\operatorname{I}}

\newcommand{\pos}[2]{\makebox(0,0)[#1]{#2}}

\newcounter{examplecntr}
{\begin{trivlist}\small\item[]\refstepcounter{examplecntr}%
 {\bfseries Example~\theexamplecntr%
  \ifthenelse{\equal{#1}{}}{}{ (#1)}.
}}%
{\end{trivlist}}

\newcounter{propositioncntr}
\newenvironment{proposition}[1][]%
{\begin{trivlist}\item[]\refstepcounter{propositioncntr}%
{\bfseries Proposition~\thepropositioncntr%
  \ifthenelse{\equal{#1}{}}{}{ (#1)}.
}}%
{\hfill\end{trivlist}}

\newcounter{remarkcntr}
\newenvironment{remark}[1][]%
{\begin{trivlist}\item[]\refstepcounter{remarkcntr}%
{\bfseries Remark~\theremarkcntr%
  \ifthenelse{\equal{#1}{}}{}{ (#1)}.
}}%
{\end{trivlist}}

\newcounter{theoremcntr}
{\begin{trivlist}\item[]\refstepcounter{theoremcntr}%
{\bfseries Theorem~\thetheoremcntr%
  \ifthenelse{\equal{#1}{}}{}{ (#1)}.
}}%
{\hfill$\Box$\end{trivlist}}

\begin{document}

%

%

\twocolumn[

\aistatstitle{Marginal Densities, Factor Graph Duality, and\\ High-Temperature Series Expansions}

\aistatsauthor{Mehdi Molkaraie}

\aistatsaddress{Department of Statistical Sciences\\ University of Toronto} ]

\begin{abstract}
We prove that the marginal densities of a global probability mass function in a
primal normal factor graph and the corresponding marginal densities in the dual normal factor 
graph are related via local mappings. The mapping depends on the Fourier transform
of the local factors of the models. Details of the mapping, including its fixed points, are derived for the 
Ising model, and then extended to the Potts model. 
By employing the mapping, we can transform 
\emph{simultaneously} all the estimated marginal densities from one domain to the other, which 
is advantageous if estimating the marginals 
can be carried out more efficiently in the dual domain.
An example of particular significance is the ferromagnetic Ising model in a positive 
external field, for which there is a rapidly mixing Markov chain (called the subgraphs-world process) 
to generate configurations in the dual normal factor graph of the model.
Our numerical experiments illustrate that the proposed procedure can provide more accurate 
estimates of marginal 
densities in various settings.

\end{abstract}

\section{Introduction}

In any probabilistic inference problem, one of the main objectives is to
compute the local marginal 
densities of a global probability mass function (PMF). Such a 
computation in general require a summation with an exponential number of terms, which makes its exact computation intractable~\cite{dagum1993approximating}. 

Our approach for estimating marginal densities hinges on the notions of the normal 
realization (in which there is an edge for every variable)~\cite{Forney:01}, the
normal factor graph (NFG), and the dual NFG. The NFG duality theorem states that the partition 
function of a primal
NFG and the partition function of its dual are equal up to some 
known scale factor~\cite{AY:2011, Forney:11}.  
It has been demonstrated that, in the low-temperature regime, Monte 
Carlo methods for estimating the partition function converge faster in the dual NFG than in the primal 
NFG of the two-dimensional (2D) 
Ising model~\cite{MoLo:ISIT2013} and of the $q$-state 
Potts model \cite{ AY:2014, MoGo:2018}.

In this paper, we prove that marginal densities of a global PMF of a primal NFG and
the corresponding marginals of the dual NFG are related via local mappings. 
Remarkably, the mapping is independent of the size of the model, of the topology of the graph, and of any assumptions on 
the parameters of the model. 

Each marginal density can of course be expressed as a ratio of two 
partition functions. In non-homogeneous models, 
each ratio needs to 
be estimated \emph{separately} via
variational inference algorithms or via Monte Carlo methods.
However, our proposed mapping allows a \emph{simultaneous} transformation of 
estimated marginal densities from one domain to the other. 

The mapping is
practically advantageous if computing such estimates
can be done more efficiently in the dual NFG than in the primal NFG.
Indeed, for the ferromagnetic Ising model in a positive 
external field there is a rapidly mixing Markov chain (called the subgraphs-world process) 
to generate configurations in the dual NFG of the Ising model.
As models, we mainly focus on binary models 
with symmetric pairwise interactions (e.g., the Ising model). However, we will briefly discuss 
extensions of the proposed mappings to non-binary models (e.g., the $q$-state Potts model).

Next, we will describe our models in the primal and in the dual domains.

%


\section{THE PRIMAL MODEL}
\label{sec:Ising} 

Suppose variables $X_1, X_2, \ldots, X_N$ are associated 
with the vertices (sites) of a connected graph 
$\G = (\VV, \EE)$ with $|\VV| = N$ vertices and $|\EE|$ edges (bonds). 
Two variables $(X_i, X_j)$ interact 
if their corresponding vertices are connected by an edge in $\G$. Each variable takes values 
in $\calA = \ZZ/2\ZZ$, i.e., the set of integers modulo two.
We will mainly view $\calA$ as a group with respect to addition.

In the primal domain, the probability of a configuration $\x \in \calA^N$ is
given by
\begin{equation} 
\label{eqn:ProbB}
\mathrm{\pi}(\x) \propto \prod_{(i,j) \in \EE} \psi_{i,j}(x_i, x_j)\prod_{v \in \VV} \phi_v(x_v). 
\end{equation}
Furthermore, we assume that each pairwise potential factor
$\psi_{i,j}(\cdot)$ is only a function of $y_{i,j} = x_i - x_j$. To lighten notations we denote the index pair
$(i, j) \in \VV^2$ by a single index $e \in \EE$. 
In the primal domain, we express the global probability 
mass function (PMF) as
\begin{equation} 
\label{eqn:ProbP}
\mathrm{\pi}_\text{p} (\x) = \frac{1}{Z_\text{p}}\prod_{e \in \EE} \psi_e(y_e)\prod_{v \in \VV} \phi_v(x_v).
\end{equation}
Here, the normalization constant $Z_\text{p}$ is the partition function, 
$\{\psi_{e} \colon \calA \rightarrow \R_{\ge 0}, e \in \EE\}$ are the \emph{edge-weighing factors}, and  
$\{\phi_{v} \colon \calA \rightarrow \R_{\ge 0}, v \in \VV\}$ are the 
\emph{vertex-weighing factors}~\cite{Mo:Allerton17, Forney:18}.

The factorization in~(\ref{eqn:ProbP}) can be represented
by an NFG $\G=(\VV,\EE)$, where vertices represent the factors 
and edges 
represent the variables.
The edge 
that represents some variable $y_e$ is connected to the vertex 
representing the factor $\psi_e(\cdot)$
if and only if $y_e$ is an argument of $\psi_e(\cdot)$. If a 
variable appears in more than two factors, it is 
replicated using an equality indicator factor~\cite{Forney:01}. 

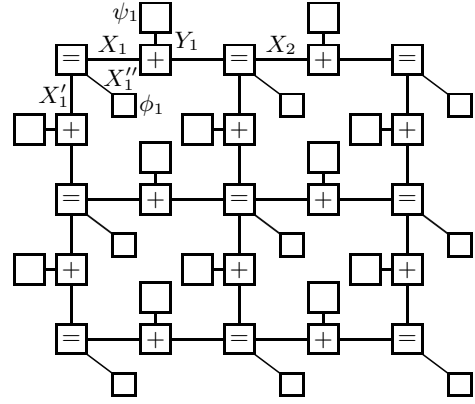
\begin{figure}[t]
\setlength{\unitlength}{0.93mm}
 \linethickness{0.22mm}
\centering
\begin{picture}(78,51)(-10,0)
\small
%
%
%
\put(0,40){\framebox(4,4){$=$}}
 \put(4,40){\line(4,-3){4}}
 \put(8,34){\framebox(3,3){}}
\put(4,42){\line(1,0){8}}
\put(12,40){\framebox(4,4){$+$}}
\put(16,42){\line(1,0){8}}
\put(24,40){\framebox(4,4){$=$}}
 \put(28,40){\line(4,-3){4}}
 \put(32,34){\framebox(3,3){}}
\put(28,42){\line(1,0){8}}
\put(36,40){\framebox(4,4){$+$}}
\put(40,42){\line(1,0){8}}
\put(48,40){\framebox(4,4){$=$}}
 \put(52,40){\line(4,-3){4}}
 \put(56,34){\framebox(3,3){}}
%
\put(2,34){\line(0,1){6}}
\put(0,30){\framebox(4,4){$+$}}
\put(2,30){\line(0,-1){6}}
\put(26,34){\line(0,1){6}}
\put(24,30){\framebox(4,4){$+$}}
\put(26,30){\line(0,-1){6}}
\put(50,34){\line(0,1){6}}
\put(48,30){\framebox(4,4){$+$}}
\put(50,30){\line(0,-1){6}}
%
\put(0,20){\framebox(4,4){$=$}}
 \put(4,20){\line(4,-3){4}}
 \put(8,14){\framebox(3,3){}}
\put(4,22){\line(1,0){8}}
\put(12,20){\framebox(4,4){$+$}}
\put(16,22){\line(1,0){8}}
\put(24,20){\framebox(4,4){$=$}}
 \put(28,20){\line(4,-3){4}}
 \put(32,14){\framebox(3,3){}}
\put(28,22){\line(1,0){8}}
\put(36,20){\framebox(4,4){$+$}}
\put(40,22){\line(1,0){8}}
\put(48,20){\framebox(4,4){$=$}}
 \put(52,20){\line(4,-3){4}}
 \put(56,14){\framebox(3,3){}}
%
\put(2,14){\line(0,1){6}}
\put(0,10){\framebox(4,4){$+$}}
\put(2,10){\line(0,-1){6}}
\put(26,14){\line(0,1){6}}
\put(24,10){\framebox(4,4){$+$}}
\put(26,10){\line(0,-1){6}}
\put(50,14){\line(0,1){6}}
\put(48,10){\framebox(4,4){$+$}}
\put(50,10){\line(0,-1){6}}
%
\put(0,0){\framebox(4,4){$=$}}
 \put(4,0){\line(4,-3){4}}
 \put(8,-6){\framebox(3,3){}}
\put(4,2){\line(1,0){8}}
\put(12,0){\framebox(4,4){$+$}}
\put(16,2){\line(1,0){8}}
\put(24,0){\framebox(4,4){$=$}}
 \put(28,0){\line(4,-3){4}}
\put(32,-6){\framebox(3,3){}}
\put(28,2){\line(1,0){8}}
\put(36,0){\framebox(4,4){$+$}}
\put(40,2){\line(1,0){8}}
\put(48,0){\framebox(4,4){$=$}}
 \put(52,0){\line(4,-3){4}}
 \put(56,-6){\framebox(3,3){}}

\put(8,42.8){\pos{bc}{$X_1$}}
\put(-0.5,35.2){\pos{bc}{$X'_1$}}
\put(9.1,37.7){\pos{bc}{$X''_1$}}
\put(32,42.8){\pos{bc}{$ X_2$}}
\put(18.5,43.4){\pos{bc}{$Y_1$}}
\put(9.8,47.2){\pos{bc}{$\psi_1$}}
\put(13.5,34){\pos{bc}{$\phi_1$}}
%
\put(14,44){\line(0,1){2}}
\put(38,44){\line(0,1){2}}
\put(12,46){\framebox(4,4){$$}}
\put(36,46){\framebox(4,4){$$}}
%
\put(14,24){\line(0,1){2}}
\put(38,24){\line(0,1){2}}
\put(12,26){\framebox(4,4){$$}}
\put(36,26){\framebox(4,4){$$}}
%
%
\put(14,4){\line(0,1){2}}
\put(38,4){\line(0,1){2}}
\put(12,6){\framebox(4,4){$$}}
\put(36,6){\framebox(4,4){$$}}
%
%
\put(0,32){\line(-1,0){2}}
\put(24,32){\line(-1,0){2}}
\put(48,32){\line(-1,0){2}}
\put(-6,30){\framebox(4,4){$$}}
\put(18,30){\framebox(4,4){$$}}
\put(42,30){\framebox(4,4){$$}}
%
%
\put(0,12){\line(-1,0){2}}
\put(24,12){\line(-1,0){2}}
\put(48,12){\line(-1,0){2}}
%
\put(-6,10){\framebox(4,4){$$}}
\put(18,10){\framebox(4,4){$$}}
\put(42,10){\framebox(4,4){$$}}
%
\end{picture}
\vspace{2.0ex}
\caption{\label{fig:2DGridMod}%
Primal NFG of the factorization~(\ref{eqn:ProbP}).
}
\end{figure}

For a 2D lattice, the NFG of~(\ref{eqn:ProbP}) is depicted in~\Fig{fig:2DGridMod},
in which the unlabeled 
boxes represent $\psi_e(\cdot)$, small unlabeled 
boxes represent $\phi_v(\cdot)$. 
In~\Fig{fig:2DGridMod}, boxes labeled ``$+$'' are instances of zero-sum indicator 
factors $\I_{+}(\cdot)$, which impose the
constraint that all their incident variables sum 
to zero, and boxes labeled ``$=$'' are instances of equality 
indicator factors $\I_{=}(\cdot)$, which impose the
constraint that all their incident variables are equal.

E.g., the equality indicator factor involving $x_1, x'_1,$ and $x''_1$ is given by 
\begin{equation}
\label{eqn:equality}
\I_{=}(x_1, x'_1, x''_1) = \delta( x_1 - x'_1)\cdot\delta(x_1- x''_1)
\end{equation}
and the zero-sum indicator factor involving 
$x_1, x_2,$ and $y_1$ is as in 
\begin{equation}
\label{eqn:XOR}
\I_{+}(y_1, x_1, x_2) = \delta (y_1 + x_1 + x_2),
\end{equation}
where $\delta(\cdot)$ is the Kronecker delta function.
(Note that all arithmetic manipulations are modulo two.)


In the primal NFG,  variables include $\X = \{X_v \colon v \in \VV \}$ and
\mbox{$\Y = \{Y_e\colon e\in \EE \}$}.
However, these variables are not independent. 
Indeed, we can freely choose $\X$ 
and therefrom fully determine $\Y$~\cite{MoGo:2018, Forney:18}. E.g., if we 
take $\G$ to be a $d$-dimensional lattice, we can compute each component $Y_e$ of $\Y$ 
by adding two components of $\X$ that are incident to the corresponding zero-sum indicator 
factor (see~\Fig{fig:2DGridMod}). 

The number of configurations in the primal domain is thus $|\calA|^N$, and
\begin{equation} 
\label{eqn:ZP}
Z_\text{p} = \sum_{\x \in \calA^N}\prod_{e \in \EE} \psi_e(y_e)\prod_{v \in \VV} \phi_v(x_v).
\end{equation}
The Ising model can be easily formulated via (\ref{eqn:ProbP}). In an Ising model the energy of a configuration $\x$ is given by the Hamiltonian\footnote{In the bipolar case (i.e., when $\calX = \{-1,+1\}$), 
the Hamiltonian is $\HH(\x) = -\sum_{(i,j) \in \EE} J_{i,j}x_ix_j - \sum_{1 \le i \le N}H_{i}x_i$.}
\begin{multline}   \label{eqn:HamiltonianIsingPair}
\HH(\x)  = -\sum_{(i,j) \in \EE} J_{i,j}\cdot\big(2\delta(x_i - x_j)-1\big) - \\[-1mm]
\sum_{v \in \VV}H_{v}\cdot\big(2\delta(x_v)-1\big),
\end{multline}
which can be expressed as
\begin{multline} \label{eqn:HamiltonianIsingPairY}
\HH(\x) = -\sum_{e \in \EE} J_{e}\cdot\big(2\delta(y_e)-1\big) - \\[-1mm]
\sum_{v \in \VV}H_{v}\cdot\big(2\delta(x_v)-1\big).
\end{multline}
Here $J_e$ is the coupling parameter 
associated with the bond $e \in \EE$ and $H_v$ is the external field at site $v \in \VV$.
The model is called \emph{homogeneous} if couplings are constant and \emph{ferromagnetic} if $J_{e} \ge 0$ for all $e \in \EE$.

The probability of  $\x$ is
given by the Gibbs-Boltzmann distribution~\cite{Yeo:92}
\begin{equation}  
\label{eqn:PBoltz}
\pi_{\text{B}}(\x) \propto e^{-\beta\HH(\x)},
\end{equation}
where $\beta \in \R_{\ge 0}$ denotes the inverse temperature.

From~(\ref{eqn:HamiltonianIsingPairY}) and (\ref{eqn:PBoltz}), it is 
straightforward to obtain the
edge-weighing factors of the Ising model as
\begin{equation} 
\label{eqn:IsingJmod}
\psi_e(y_e) = \left\{ \begin{array}{ll}
    e^{\beta J_e}, & \text{if $y_e = 0$} \\
     e^{-\beta J_e}, & \text{if $y_e = 1$}
  \end{array} \right.
\end{equation}
and the vertex-weighing factors as
\begin{equation} 
\label{eqn:IsingH}
\phi_{v}(x_v) = \left\{ \begin{array}{ll}
     e^{\beta H_{v}}, & \text{if $x_v = 0$} \\
     e^{-\beta H_{v}}, & \text{if $x_v = 1$.}
  \end{array} \right.
\end{equation}
The Gibbs-Boltzmann distribution in (\ref{eqn:PBoltz}) can therefore be expressed via the 
factorization~(\ref{eqn:ProbP}).

\section{THE DUAL MODEL}
\label{sec:IsingD} 

The dual NFG has the same topology as the primal NFG, but  with
factors replaced by the discrete Fourier transform (DFT) or the inverse DFT 
of corresponding 
factors in the primal NFG.

We can obtain the dual NFG of our binary models by
replacing factors by their one-dimensional (1D) DFT, equality indicator 
factors by zero-sum indicator factors, and zero-sum indicator factors by 
equality indicator factors~\cite{AY:2011, MoLo:ISIT2013, Mo:IZS2016}. 

We will use the tilde symbol to denote variables in the dual NFG,
which also take values in $\calA$.

\begin{figure}[t]
\setlength{\unitlength}{0.93mm}
 \linethickness{0.22mm}
\centering
\begin{picture}(78,52)(-10,0)
\small
%
%
%
\put(0,40){\framebox(4,4){$+$}}
 \put(4,40){\line(4,-3){4}}
 \put(8,34){\framebox(3,3){}}
\put(4,42){\line(1,0){8}}
\put(12,40){\framebox(4,4){$=$}}
\put(16,42){\line(1,0){8}}
\put(24,40){\framebox(4,4){$+$}}
 \put(28,40){\line(4,-3){4}}
 \put(32,34){\framebox(3,3){}}
\put(28,42){\line(1,0){8}}
\put(36,40){\framebox(4,4){$=$}}
\put(40,42){\line(1,0){8}}
\put(48,40){\framebox(4,4){$+$}}
 \put(52,40){\line(4,-3){4}}
 \put(56,34){\framebox(3,3){}}
%
\put(2,34){\line(0,1){6}}
\put(0,30){\framebox(4,4){$=$}}
\put(2,30){\line(0,-1){6}}
\put(26,34){\line(0,1){6}}
\put(24,30){\framebox(4,4){$=$}}
\put(26,30){\line(0,-1){6}}
\put(50,34){\line(0,1){6}}
\put(48,30){\framebox(4,4){$=$}}
\put(50,30){\line(0,-1){6}}
%
\put(0,20){\framebox(4,4){$+$}}
 \put(4,20){\line(4,-3){4}}
 \put(8,14){\framebox(3,3){}}
\put(4,22){\line(1,0){8}}
\put(12,20){\framebox(4,4){$=$}}
\put(16,22){\line(1,0){8}}
\put(24,20){\framebox(4,4){$+$}}
 \put(28,20){\line(4,-3){4}}
 \put(32,14){\framebox(3,3){}}
\put(28,22){\line(1,0){8}}
\put(36,20){\framebox(4,4){$=$}}
\put(40,22){\line(1,0){8}}
\put(48,20){\framebox(4,4){$+$}}
 \put(52,20){\line(4,-3){4}}
 \put(56,14){\framebox(3,3){}}
%
\put(2,14){\line(0,1){6}}
\put(0,10){\framebox(4,4){$=$}}
\put(2,10){\line(0,-1){6}}
\put(26,14){\line(0,1){6}}
\put(24,10){\framebox(4,4){$=$}}
\put(26,10){\line(0,-1){6}}
\put(50,14){\line(0,1){6}}
\put(48,10){\framebox(4,4){$=$}}
\put(50,10){\line(0,-1){6}}
%
\put(0,0){\framebox(4,4){$+$}}
 \put(4,0){\line(4,-3){4}}
 \put(8,-6){\framebox(3,3){}}
\put(4,2){\line(1,0){8}}
\put(12,0){\framebox(4,4){$=$}}
\put(16,2){\line(1,0){8}}
\put(24,0){\framebox(4,4){$+$}}
 \put(28,0){\line(4,-3){4}}
\put(32,-6){\framebox(3,3){}}
\put(28,2){\line(1,0){8}}
\put(36,0){\framebox(4,4){$=$}}
\put(40,2){\line(1,0){8}}
\put(48,0){\framebox(4,4){$+$}}
 \put(52,0){\line(4,-3){4}}
 \put(56,-6){\framebox(3,3){}}
\put(8,42.8){\pos{bc}{$\tilde X_1$}}
\put(32,42.8){\pos{bc}{$\tilde X_2$}}
\put(-0.8,35.3){\pos{bc}{$\tilde X_3$}}
\put(9.0,37.9){\pos{bc}{$\tilde Y_1$}}
\put(19,46.8){\pos{bc}{$\tilde \psi_1$}}
\put(13.5,34){\pos{bc}{$\tilde \phi_1$}}
%
\put(14,44){\line(0,1){2}}
\put(38,44){\line(0,1){2}}
\put(12,46){\framebox(4,4){$$}}
\put(36,46){\framebox(4,4){$$}}
%
\put(14,24){\line(0,1){2}}
\put(38,24){\line(0,1){2}}
\put(12,26){\framebox(4,4){$$}}
\put(36,26){\framebox(4,4){$$}}
%
%
\put(14,4){\line(0,1){2}}
\put(38,4){\line(0,1){2}}
\put(12,6){\framebox(4,4){$$}}
\put(36,6){\framebox(4,4){$$}}
%
%
\put(0,32){\line(-1,0){2}}
\put(24,32){\line(-1,0){2}}
\put(48,32){\line(-1,0){2}}
\put(-6,30){\framebox(4,4){$$}}
\put(18,30){\framebox(4,4){$$}}
\put(42,30){\framebox(4,4){$$}}
%
%
\put(0,12){\line(-1,0){2}}
\put(24,12){\line(-1,0){2}}
\put(48,12){\line(-1,0){2}}
%
\put(-6,10){\framebox(4,4){$$}}
\put(18,10){\framebox(4,4){$$}}
\put(42,10){\framebox(4,4){$$}}
%
\end{picture}
\vspace{2.0ex}
\caption{\label{fig:2DGridModD}%
The dual of the NFG in \Fig{fig:2DGridMod}. 
}
\end{figure}
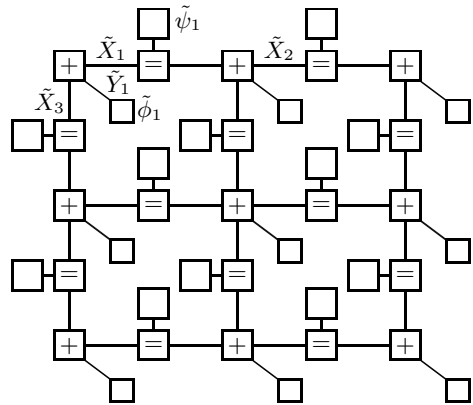

The dual NFG of~\Fig{fig:2DGridMod} is illustrated in~\Fig{fig:2DGridModD}, in 
which the 
unlabeled boxes represent 
$\tilde \psi_{e} \colon \calA \rightarrow \R$, the 1D DFT 
of $\psi_{e}(\cdot)$, given by
\begin{equation} 
\label{eqn:EdgeD}
\tilde \psi_{e}(\tilde y_e) = \left\{ \begin{array}{ll}
      \psi_{e}(0) + \psi_{e}(1), & \text{if $\tilde y_e = 0$} \\
      \psi_{e}(0) - \psi_{e}(1), & \text{if $\tilde y_e = 1$}
  \end{array} \right.
\end{equation}
and for $v \in \VV$ the small 
unlabeled boxes are $\tilde \phi_{v}\colon \calA \rightarrow \R$, the 
1D DFT of $\phi_{v}(\cdot)$, as in
\begin{equation} 
\label{eqn:NodeD}
\tilde \phi_{v}(\tilde x_v) = \left\{ \begin{array}{ll}
      \phi_{v}(0) + \phi_{v}(1), & \text{if $\tilde x_v = 0$} \\
      \phi_{v}(0) - \phi_{v}(1), & \text{if $\tilde x_v = 1$.}
  \end{array} \right.
\end{equation}

The set of variables in the dual domain consist of
\mbox{$\tilde\Y = \{ \tilde Y_e\colon e\in \EE \}$} 
and $\tilde\X = \{\tilde X_v \colon v \in \VV \}$.
Again, these variables are not independent as we can freely choose $\tilde\Y$ and 
therefrom fully determine $\tilde \X$.
E.g., if we take $\G$ to be a $d$-dimensional lattice 
and assume periodic boundaries, each component $\tilde X_v$ of $\tilde \X$  can be computed 
by adding $2d$ components of $\tilde \Y$ that are incident to the corresponding zero-sum 
indicator factor (see~\Fig{fig:2DGridModD}).

In the dual NFG, the number of configurations is $|\calA|^{|\EE|}$, and 
its the partition function $Z_\text{d}$ is given by
\begin{equation}
\label{eqn:Zd}
Z_\text{d} = \sum_{\tilde \y \in \calA^{|\EE|}} \prod_{e \in \EE} \tilde \psi_e(\tilde y_e) \prod_{v \in \VV} \tilde \phi_v(\tilde x_v).
\end{equation} 
On condition that factors (\ref{eqn:EdgeD}) and (\ref{eqn:NodeD}) are nonnegative, we can define
the global PMF in the dual NFG as
\begin{equation}
\label{eqn:Pd}
\mathrm{\pi}_\text{d}(\tilde \y) = \frac{1}{Z_\text{d}}\prod_{e \in \EE} \tilde \psi_e(\tilde y_e) \prod_{v \in \VV} \tilde \phi_v(\tilde x_v).
\end{equation} 

The dual Ising model can be expressed via (\ref{eqn:Pd}). Indeed
\begin{equation} 
\label{eqn:IsingDJ}
\tilde \psi_{e}(\tilde y_e) = \left\{ \begin{array}{ll}
      2\cosh(\beta J_e), & \text{if $\tilde y_e = 0$} \\
      2\sinh(\beta J_e), & \text{if $\tilde y_e = 1,$}
  \end{array} \right.
\end{equation}
in agreement with (\ref{eqn:IsingJmod}) and (\ref{eqn:EdgeD}), and
\begin{equation} 
\label{eqn:IsingDH}
\tilde \phi_{v}(\tilde x_v) = \left\{ \begin{array}{ll}
      2\cosh(\beta H_v), & \text{if $\tilde x_v = 0$} \\
      2\sinh(\beta H_v), & \text{if $\tilde x_v = 1$,}
  \end{array} \right.
\end{equation}
in agreement with (\ref{eqn:IsingH}) and (\ref{eqn:NodeD}). 

If the model is ferromagnetic 
(i.e., $J_e \ge 0$ ) and in a nonnegative external field (i.e., $H_v \ge 0$), 
factors (\ref{eqn:IsingDJ}) and (\ref{eqn:IsingDH}) will be nonnegative. 
In this case, the global PMF of the dual Ising model is given by~(\ref{eqn:Pd}).

Throughout this paper, we assume that each edge connects two distinct vertices of the NFG (i.e., there are no dangling edges
with one end attached to a vertex and the other end free).
In this setting, according to the NFG duality
theorem, the partition functions $Z_\text{p}$ and $Z_\text{d}$ are equal up to some scale 
factor $\alpha(\G)$. Indeed
\begin{equation}
\label{eqn:scaleF}
Z_{\text{d}} = \alpha(\G)\cdot Z_\text{p},
\end{equation} 
where $\alpha(\G)$ only depends on the topology of $\G$.

For more details, see~\cite{AY:2011}, \cite[Appendix]{Mo:Allerton17},\cite[Thm 8]{Forney:18}. 

\section{THE DUAL ISING MODEL AND 
HIGH-TEMPERATURE SERIES EXPANSIONS}

In~\cite{JS:93}, the authors proposed a rapidly mixing Markov chain (called the 
subgraphs-world process) which evaluates the partition function of an
arbitrary ferromagnetic Ising model in a positive external field to any specified 
degree of accuracy. 

The mixing time of the process is polynomial in the size of the model 
at \emph{all} temperatures. Indeed, the expected running time of the generator 
of the subgraphs-world process is $\mathcal{O}\left(|\EE|^2N^8(\log\delta^{-1} + |\EE|)\right)$, 
where $\delta$ is the confidence parameter.
For more details, see~\cite[Section 4]{JS:93}. 

The subgraphs-world process employs the following expansion of $Z$ 
defined on the set of edges $\calW \subseteq \EE$
in powers of $\tanh(H)$ and $\tanh(J_e)$ as
\begin{equation} 
\label{eqn:HTExpandExt}
Z \propto 
\sum_{\calW \subseteq \EE}\tanh(H)^{|\text{odd$(\calW)$}|}\prod_{e \in \calW} \tanh(J_e),
\end{equation}
where $\text{odd$(\calW)$}$ denotes the set of all odd-degree vertices in the subgraph of $\EE$
induced by $\calW$. 
The expansion~(\ref{eqn:HTExpandExt}) is known as 
the high-temperature series expansion of the partition 
function \cite{Newell:53, Yeo:92, grimmett:09}.

\begin{proposition} \label{prop:Hightemp}
The configurations that arise in the high-temperature series expansion of the
partition function (which 
are the configurations of the subgraphs-world process) coincide
with the valid configurations in the dual NFG of the Ising model. 
\end{proposition}
See~\cite[Section VIII]{MoGo:2018} and \cite[Section III-E]{Forney:18} for the proof.

Following Proposition~\ref{prop:Hightemp}, we can employ the subgraphs-world 
process (as a generator for the subgraphs-world configurations) to generate configurations 
in the dual NFG of the Ising model. The process is rapidly mixing and therefore converges
in polynomial time. However, under reasonable complexity assumptions, there is no
generalization of this approximation scheme to the (nonbinary) Potts model 
or to spin glasses.  For more details, see~\cite{goldberg2012, galanis2016ferromagnetic}.

Next, we will present local (edge-based) mappings that transform marginal densities from the dual NFG
to the primal NFG, or vice versa. The mappings depend on the DFT 
of the local factors of the models.

\section{MARGINAL DENSITIES IN THE PRIMAL AND DUAL DOMAINS}
\label{sec:Marg}

The edge marginal PMF of $e \in \EE$ in the primal NFG can be computed as
\begin{equation}
\label{eqn:MargProbP1}
\pi_{\text{p}, e}(a)  = \frac{Z_{\text{p}, e}(a)}{Z_\text{p}}, \quad a \in \calA,
\end{equation}
where
\begin{equation*}
Z_{\text{p}, e}(a) = \sum_{\x \in \calA^N} \delta(y_e - a)\prod_{\e \in \EE} \psi_\e(y_\e) \prod_{v \in \VV}\phi_v(x_v).
\label{eqn:MargProbUD}
\end{equation*}
Hence
\begin{equation}
Z_{\text{p}, e}(a) = \psi_e(a) S_e(a),
\label{eqn:MargProbUD}
\end{equation}
with
\begin{equation}
S_e(a) = \sum_{\x \in \calA^N} \delta(y_e - a)\prod_{\e \in \EE \setminus e} \psi_\e(y_\e) \prod_{v \in \VV}\phi_v(x_v).
\label{eqn:MargProbUD}
\end{equation}
Here, $Z_{\text{p}, e}(a) \ge 0$ and $Z_\text{p} = \sum_{a \in \calA} Z_{\text{p}, e}(a) = \sum_{a \in \calA}\psi_e(a) S_e(a)$, hence (\ref{eqn:MargProbP1}) is a valid PMF over $\calA$.

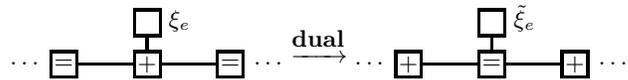
\begin{figure}[t]
\setlength{\unitlength}{0.79mm}
\centering
\begin{picture}(103,13.2)(0,0)
 \linethickness{0.27mm}
\small
\put(0,2){$\ldots$}
\put(7,0){\framebox(4,4){$=$}}
\put(11,2){\line(1,0){10}}
\put(21,0){\framebox(4,4){$+$}}
\put(25,2){\line(1,0){10}}
\put(35,0){\framebox(4,4){$=$}}
\put(41,2){$\ldots$}
\put(23,4){\line(0,1){3}}
\put(21,7){\framebox(4,4){}}
\put(52,2.0){\pos{bc}{$\parbox{.75cm}{\rightarrowfill}$}}
\put(51.8,5.0){\pos{bc}{{\bf dual}}}
\put(58,2){$\ldots$}
\put(65,0){\framebox(4,4){$+$}}
\put(69,2){\line(1,0){10}}
\put(79,0){\framebox(4,4){$=$}}
\put(83,2){\line(1,0){10}}
\put(93,0){\framebox(4,4){$+$}}
\put(99,2){$\ldots$}
\put(81,4){\line(0,1){3}}
\put(79,7){\framebox(4,4){}}
%
 \put(28.5,7.5){\pos{bc}{$\xi_e$}}
 \put(86.5,7.5){\pos{bc}{$\tilde \xi_e$}}
\end{picture}
\vspace{0.5ex}
\caption{\label{fig:Marg2}%
The edge $e \in \EE$ in the intermediate primal NFG (left) and in the intermediate dual  NFG (right).
The unlabeled box (left) represents~(\ref{eqn:Int1}) and the unlabeled box (right) represents~(\ref{eqn:Int2}).}
\end{figure}


In coding theory terminology, $\{\psi_e(a), a \in \calA \}$ is called the \emph{intrinsic} message 
vector and $\{ S_e(a), a \in \calA \}$ is called the \emph{extrinsic} message vector at edge $e \in \EE$. 
According to the sum-product message passing update rule, the edge marginal PMF vector is computed as the dot 
product of the intrinsic and extrinsic message vectors 
up to scale. The scale factor is equal to the partition 
function $Z_{\text{p}}$~\cite{Forney:01, KFL:01}. 
 
In our setup, $S_e(a)$ is the partition function of 
an intermediate primal NFG with all factors as in the primal NFG, excluding 
the factor $\psi_e(y_e)$, which is replaced by
\begin{equation}
\label{eqn:Int1}
\xi_e(y_e; a) = \delta(y_e - a).
\end{equation}
\Fig{fig:Marg2} (left) shows the corresponding edge in the intermediate primal NFG. The intermediate dual NFG is shown in~\Fig{fig:Marg2} (right), in which the factor $\tilde \psi_e(\tilde y_e)$ is replaced by
\begin{equation}
\label{eqn:Int2}
\tilde \xi_e(\tilde y_e; a) =
 \begin{cases}
   \delta(a) + \delta(1-a), & \text{if $\tilde y_e=0$}  \vspace{2pt}\\ 
   \delta(a) -  \delta(1-a), & \text{if $\tilde y_e = 1,$}
 \end{cases}
\end{equation}
which is the 1D DFT of~(\ref{eqn:Int1}). According to the NFG duality theorem~(\ref{eqn:scaleF}), 
the partition function of the intermediate dual NFG is $\alpha(\G)\cdot S_e(a)$. 

Similarly, in the dual NFG the edge marginal PMF of $e \in \EE$ is
\begin{equation}
\label{eqn:MargProbD}
\pi_{\text{d}, e}(a')  = \frac{Z_{\text{d}, e}(a')}{Z_{\text{d}}}, \quad a' \in \calA .
\end{equation}
Hence
\begin{align}
Z_{\text{d}, e}(a')
             & = \tilde \psi_e(a')\cdot \notag \\
             & \Big(\sum_{\tilde \y \in \calA^{|\EE|}} \delta(\tilde y_e - a')
             \prod_{\e \in \EE \setminus e} \tilde \psi_\e(\tilde y_\e) \prod_{v \in \VV}\tilde \phi_v(\tilde x_v)\Big) \notag \\
            & = \tilde \psi_e(a') \tilde S_e(a'). \label{eqn:MargProbUD2}
\end{align}

\begin{proposition} \label{prop:cutset}
The vectors $\{ S_e(a), a\in \calA \}$ and 
$\{ \tilde S_e(a'), a'\in \calA \}$ are DFT pairs.
\end{proposition}

\begin{trivlist}
\item \emph{Proof. } 
For $a \in \calA$, the partition function of the intermediate dual NFG is the dot product of 
message vectors $\{ \tilde \xi_e(a'; a), a'\in \calA \}$ and $\{ \tilde S_e(a'), a' \in \calA \}$.
Thus
\begin{equation}
\alpha(\G)\cdot S_e(a) = \sum_{a' \in \calA} \tilde \xi_e(a'; a)\tilde S_e(a'),
\end{equation}
which gives
\begin{multline}
\alpha(\G)\cdot S_e(a)   
             = \big(\tilde S_e(0) + \tilde S_e(1)\big)\cdot\delta(a) + \\
             \big(\tilde S_e(0)-\tilde S_e(1)\big)\cdot\delta(1-a). \label{eqn:InterimNFGD}
\end{multline}

After setting $a = 0$ and $a = 1$ in~(\ref{eqn:InterimNFGD}), we obtain
\begin{equation}
\label{eqn:InterimZDuality}
\begin{bmatrix} \, S_e(0)\,  \\[6pt] \, S_e(1)\, \end{bmatrix} = \frac{1}{\alpha(\G)}
\begingroup
\renewcommand*{\arraystretch}{1.0}
\begin{bmatrix}      
\, 1 & 1\, \\[6pt] \,1 & - 1\,
\end{bmatrix}
\endgroup
\cdot \begin{bmatrix} \, \tilde S_e(0) \, \\[6pt] \, \tilde S_e(1)\, \end{bmatrix}
\end{equation}
which is an instance of the two-point DFT.
\hfill$\blacksquare$
\end{trivlist}

\begin{proposition} \label{prop:mappingDFT}
The vectors $\{ \pi_{\text{p}, e}(a)/\psi_e(a), a\in \calA \}$ and 
$\{\pi_{\text{d}, e}(a')/\tilde \psi_e(a'), a'\in \calA \}$ are DFT pairs.
\end{proposition}
\begin{trivlist}
\item \emph{Proof. } 
From (\ref{eqn:MargProbP1}) and (\ref{eqn:MargProbUD}) we have
\begin{equation}
S_e(a) = Z_{\text{p}}\cdot \frac{\pi_{\text{p}, e}(a)}{\psi_e(a)}, \quad a \in \calA. \label{eqn:MargProbInterim2} 
\end{equation}
But (\ref{eqn:scaleF}), (\ref{eqn:MargProbD}), and (\ref{eqn:MargProbUD2}) yield
\begin{align}
\tilde S_e(a') & = Z_{\text{d}}\cdot \frac{\pi_{\text{d}, e}(a')}{\tilde \psi_e(a')}\\ 
 & = \alpha(\G)\cdot Z_{\text{p}}\cdot\frac{\pi_{\text{d}, e}(a')}{\tilde \psi_e(a')} , \quad a' \in \calA. \label{eqn:MargProbInterim3}
\end{align}


Putting (\ref{eqn:MargProbInterim3}) and (\ref{eqn:MargProbInterim2}) in (\ref{eqn:InterimZDuality}), and after a little rearranging, we obtain 
the following mapping 
\begin{equation}
\label{eqn:MapG}
\begin{bmatrix} \, \pi_{\text{p}, e}(0)/\psi_e(0)\,  \\[8pt] \, \pi_{\text{p}, e}(1)/\psi_e(1)\, \end{bmatrix} = 
\begingroup
\renewcommand*{\arraystretch}{1.0}
\begin{bmatrix}      
\, 1 & 1\, \\[6pt] \,1 & - 1\,
\end{bmatrix}
\endgroup
\cdot \begin{bmatrix} \, \pi_{\text{d}, e}(0)/\tilde \psi_e(0) \, \\[8pt] \, \pi_{\text{d}, e}(1)/\tilde \psi_e(1)\, \end{bmatrix}
\end{equation}
in matrix-vector format via the two-point DFT.
\hfill$\blacksquare$
\end{trivlist}

By virtue of Proposition~\ref{prop:mappingDFT}, it is possible to estimate
edge marginal densities in one domain, and then transform them to the other domain all together.
The mapping is fully local, and is independent 
of the size of the graph $N$ and of the topology of~$\G$. 
(Indeed, the relevant information regarding the rest of the graph is incorporated in the 
estimated edge marginal densities.)

We state without proof that
\begin{proposition} \label{prop:HmappingDFT}
The vectors $\{\pi_{\text{p}, v}(a)/\phi_v(a), a \in \calA \}$ 
and $\{\pi_{\text{d}, v}(a')/\tilde \phi_v(a'), a' \in \calA \}$ are DFT pairs. 
\end{proposition}

\section{DETAILS OF THE MAPPING FOR THE ISING MODEL}
\label{sec:IsingModelMapping}


For the general Ising model substituting factors~(\ref{eqn:IsingJmod}) and~(\ref{eqn:IsingDJ}) in (\ref{eqn:MapG}) yields
\begin{equation}
\label{eqn:MapDP}
\begin{bmatrix} \, \pi_{\text{p},e}(0)\,  \\[6pt] \,\pi_{\text{p},e}(1)\, \end{bmatrix} = 
\begingroup
\renewcommand*{\arraystretch}{1.9}
\begin{bmatrix}      
\,\dfrac{e^{\beta J_e}}{2\cosh(\beta J_e)} & \dfrac{e^{\beta J_e}}{2\sinh(\beta J_e)}\, \\[8pt] \,\dfrac{e^{-\beta J_e}}{2\cosh(\beta J_e)} & - {\dfrac{e^{-\beta J_e}}{2\sinh(\beta J_e)}}\,
\end{bmatrix}
\endgroup
\cdot \begin{bmatrix} \, \pi_{\text{d},e}(0)\, \\[6pt] \, \pi_{\text{d},e}(1)\, \end{bmatrix} 
\end{equation}
for $\beta J_e \ne 0$.

Let us consider a homogeneous and ferromagnetic Ising model. 
A straightforward calculation shows that 
the fixed points of the mapping (\ref{eqn:MapDP}) are given by
\begin{multline}
\label{eqn:fixed}
\begin{bmatrix} \, \pi^{*}_{\text{p},e}(0)  & \pi^{*}_{\text{p},e}(1)\, \end{bmatrix} = \\
\begin{bmatrix} \, \dfrac{e^{\beta J}\cosh (\beta J)}{1+\sinh(2\beta J)}\, & \,\dfrac{e^{-\beta J}\sinh(\beta J)}{1+\sinh(2\beta J)}\, \end{bmatrix}
\end{multline}

Fig.~\ref{fig:L1} shows the fixed points (\ref{eqn:fixed}) as a function of $\beta J$.

\begin{proposition} \label{prop:2DHom}
The min of $\pi^{*}_{\text{p},e}(0)$ and the max of $\pi^{*}_{\text{p},e}(1)$ are attained at the criticality of the 
2D homogeneous Ising model without an external field. 
\end{proposition}

\begin{trivlist}
\item \emph{Proof. } 
In the thermodynamic limit (i.e., as $N \to \infty$) the 2D Ising model undergoes 
a phase transition at 
$\beta J_c =  \ln(1+\sqrt{2})/2 \approx 0.44$~\cite{Onsager:44}. 

In the absence of an external field and for $\beta J = 1$, 
the Hamiltonian (\ref{eqn:HamiltonianIsingPairY}) can be expressed as
\begin{align}
\label{eqn:HamP}
\mathcal{H}(\y)  & = -\sum_{e \in \EE}\big(2\delta(y_e) -1\big)\\
						  & = -\sum_{e \in \EE}(1-2y_e),
\end{align}
where $y_e = x_i - x_j$ for $e = (i, j) \in \EE$.

The average energy is equal to 
\begin{align}
\label{eqn:aveE}
\overbar{\mathcal{H}}(\y) & = \sum_{\y \in \calA}\pi_\text{p}(\y)\mathcal{H}(\y)\\ 
	& = -|\EE|\cdot(1-2\E[Y_e])\\
	& = -|\EE|\cdot(1-2\pi_{\text{p},e}(1)).
\end{align}

In the 2D Ising model with periodic boundaries $|\EE| = 2N$, thus the average energy
per site is given by
\begin{equation}
\label{eqn:AveEper}
\overbar{\mathcal{H}}(\y)/N  = -2(1-2\pi_{\text{p},e}(1)).
\end{equation}

From Onsager's closed-form solution
\begin{multline}
\label{eqn:ZOnsg}
\lim_{N \to \infty} \frac{\ln Z}{N} = \frac{1}{2}\ln(2\cosh^{2}2\beta J)  + \\
\frac{1}{\pi}\int_{0}^{\frac{\pi}{2}}\ln\big(1+\sqrt{1-\kappa^2\sin^2\theta}\,\big)\,d\theta
\end{multline}
and the average (internal) energy per site is given by
\begin{align}
\label{eqn:IntEPerSite}
U(\beta J) & = \lim_{N \to \infty} -\frac{1}{N}\cdot\frac{\partial \ln Z}{\partial \beta J} \\
& = -\coth(2\beta J)\cdot\notag \\
&\!\!\!\!\!\!\!\Big(1-\frac{1}{2\pi}(1-\kappa\sinh2\beta J)\!\int_{0}^{\frac{\pi}{2}}\!\!\!\!\frac{d\theta}{\sqrt{1-\kappa^2\sin^2\theta}}\Big)
\end{align}
with
\begin{equation}
\label{eqn:Cons}
\kappa(\beta J) = \frac{2\sinh2\beta J}{\cosh^{2}2\beta J}\cdot
\end{equation}
See~\cite{Onsager:44}, \cite[Chapter 7]{Baxter:07} for more details.

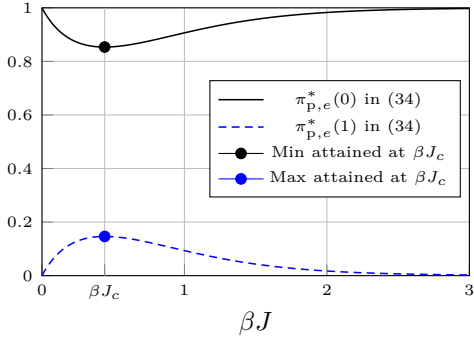
\begin{figure}[t!!]
\centering
\begin{tikzpicture}
\begin{axis}[
			legend style={at = {(0.98,0.73)} ,font=\tiny},		
			height = 34.0ex,
			width = 48ex,
			grid = major,
			tick pos=left, 
			xlabel shift = -2 pt,
			xminorticks = false,	
		    yminorticks = false,	
		    y tick label style={
        /pgf/number format/.cd,
            fixed,
        /tikz/.cd
    		}, 				
			ytick={0, 0.2, 0.4, 0.6, 0.8, 1.0},
			xtick={0.0, 1.0, 2.0, 3.0},
			extra x ticks=0.44,
			extra x tick labels={$\beta J_c$},
		xlabel= $\beta J$ ={font=\normalsize},
			xmin = 0.0,
			xmax = 3.0,
			ymin = 0.0,
			ymax = 1.0,
			yticklabel style = {font=\tiny,yshift=0.0ex},
            xticklabel style = {font=\tiny,xshift=0.0ex}			
			]

\pgfplotstableread{./Pfixed.txt}\mydataone
\pgfplotstableread{./Pfixed2.txt}\mydatatwo

		\addplot [
		line width = 0.58 pt,
		 color = black
		]		
		 table[y = Z] from \mydataone;
		 
 		 \addplot [
 		 densely dashed,
 		 line width = 0.58 pt,
 		 color = blue
 		 ]
 		  table[y = Z] from \mydatatwo;	 
 
		\addplot[mark=*] coordinates {(0.44,0.853554)};
		\addplot[color = blue, mark=*] coordinates {(0.44,0.146446)};


 		 \legend{$\pi^{*}_{\text{p},e}(0)$ in~(\ref{eqn:fixed}), $\pi^{*}_{\text{p},e}(1)$ in~(\ref{eqn:fixed}), Min attained at $\beta J_c$, 
 		 Max attained at $\beta J_c$};	  	
\end{axis}
\end{tikzpicture}
\vspace{-2.0ex}
\caption{\label{fig:L1}%
The fixed points~(\ref{eqn:fixed}) as a function of 
$\beta J$. 
The filled circles show the fixed points at criticality of the 2D Ising model given by (\ref{eqn:fixedC}).}
\end{figure}

A routine calculation shows that $\kappa(\beta J_c) =1$, thus
\begin{equation}
\label{eqn:IntEPT}
U(\beta J_c) = -\sqrt{2}.
\end{equation}
From (\ref{eqn:AveEper}) and (\ref{eqn:IntEPT}), we 
obtain $\pi^{*}_{\text{p},e}(1) = (2-\sqrt{2})/4$. Therefore, at criticality
\begin{multline}
\label{eqn:fixedC}
\begin{bmatrix} \, \pi^{*}_{\text{p},e}(0)  & \pi^{*}_{\text{p},e}(1)\, \end{bmatrix}  = \\
\begin{bmatrix} \, (2+\sqrt{2})/4\, & \, (2-\sqrt{2})/4\, \end{bmatrix}, 
\end{multline}
which coincides with the min of $\pi^{*}_{\text{p},e}(0)$ and the max
of $\pi^{*}_{\text{p},e}(1)$. We emphasize that in the 2D homogeneous Ising model in zero field and
in the thermodynamic limit (i.e., as $N \to \infty$), edge marginal densities in the primal and dual domains 
are equal at criticality.
\hfill$\blacksquare$
\end{trivlist}

The fixed points $\pi^{*}_{\text{p},e}(0)$ and $\pi^{*}_{\text{p},e}(1)$ at criticality in~(\ref{eqn:fixedC}) are 
illustrated by filled circles in Fig.~\ref{fig:L1}.

\begin{proposition} \label{prop:BoundsOnP}
In an arbitrary ferromagnetic Ising model in a nonnegative external field, it holds that
\begin{equation}
\label{eqn:BoundonPe0}
\pi_{\text{p},e}(0) \ge \frac{1}{1+e^{-2\beta J_e}}
\end{equation}
and
\begin{equation}
\label{eqn:BoundonPd0}
\pi_{\text{d},e}(0) \ge \frac{1+e^{-2\beta J_e}}{2}.
\end{equation}
\end{proposition}

\begin{trivlist}
\item \emph{Proof. } Since the Ising model is ferromagnetic and in a nonnegative external field, we can define 
the global PMF $\pi_{\text{d},e}(\cdot)$ 
in the dual domain as in~(\ref{eqn:Pd}). From~(\ref{eqn:MapDP}), we have
\begin{align}
\frac{\pi_{\text{p},e}(0)}{e^{\beta J_e}}  
& = \frac{\pi_{\text{d},e}(0)}{2\cosh(\beta J_e)} + \frac{\pi_{\text{d},e}(1)}{2\sinh(\beta J_e)}\\
& = \frac{1}{2\sinh(\beta J_e)} - \frac{e^{-\beta J_e}}{\sinh(2\beta J_e)}\pi_{\text{d},e}(0). \label{eqn:BoundonPeDerive}
\end{align}

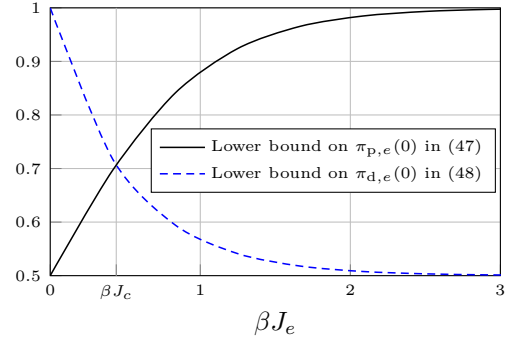
\begin{figure}[t!!]
\centering
\begin{tikzpicture}
\begin{axis}[
			legend style={at = {(0.99,0.55)} ,font=\tiny},		
			height = 34.0ex,
			width = 50ex,
			grid = major,
			tick pos=left, 
			xlabel shift = -2 pt,
			restrict y to domain = 0.5:1, 
			xminorticks = false,	
		    yminorticks = false,	
		    y tick label style={
        /pgf/number format/.cd,
            fixed,
        /tikz/.cd
    		}, 				
			ytick={0.5, 0.6, 0.7, 0.8, 0.9, 1.0},
			xtick={0.0, 1.0, 2.0, 3.0},
			extra x ticks=0.44,
			extra x tick labels={$\beta J_c$},
		xlabel= $\beta J_e$ ={font=\normalsize},
			xmin = 0.0,
			xmax = 3.0,
			ymin = 0.5,
			ymax = 1.0,
			yticklabel style = {font=\tiny,yshift=0.0ex},
            xticklabel style = {font=\tiny,xshift=0.0ex}			
			]

\addplot[black, line width = 0.58 pt, smooth]    {e^(2*x)/(1+e^(2*x))};
\addplot[blue, line width = 0.58 pt, smooth, densely dashed]    {(1+e^(2*x))/(2*e^(2*x))};

 		 \legend{Lower bound on $\pi_{\text{p},e}(0)$ in~(\ref{eqn:BoundonPe0}), Lower bound on $\pi_{\text{d},e}(0)$ in~(\ref{eqn:BoundonPd0})};	  	
\end{axis}
\end{tikzpicture}
\vspace{-2.0ex}
\caption{\label{fig:PBound}%
For a ferromagnetic Ising model in a nonnegative external field,
the solid black plot and the dashed blue plot show the lower bound on $\pi_{\text{p},e}(0)$, given by (\ref{eqn:BoundonPe0}), and 
the lower bound on $\pi_{\text{d},e}(0)$, given by (\ref{eqn:BoundonPd0}), as a function of $\beta J_e$, respectively. 
The lower bounds intersect at  the criticality of the 
2D homogeneous Ising model in zero field, denoted by $\beta J_c$.}
\end{figure}


We conclude from~(\ref{eqn:BoundonPeDerive}) that $\pi_{\text{p},e}(0)$ achieves its minimum when $ \pi_{\text{d},e}(0)= 1$. 
After substituting $\pi_{\text{d},e}(0) = 1$ in~(\ref{eqn:BoundonPeDerive}), and after a little rearranging, we obtain
\begin{align}
\pi_{\text{p},e}(0) & \ge \frac{e^{\beta J_e}}{2\cosh(\beta J_e)} \\
                             & = \frac{1}{1+e^{-2\beta J_e}}\cdot \label{eqn:BoundonProofs}
\end{align}
The proof of~(\ref{eqn:BoundonPd0}) follows along the same lines. 
\hfill$\blacksquare$
\end{trivlist}

Proposition~\ref{prop:BoundsOnP} is valid for arbitrary ferromagnetic 
Ising models in a nonnegative external magnetic field, i.e., the bonds do not depend on $N$ (the size of the 
graphical model $\G$) and on the 
topology of $\G$.
 
Fig.~\ref{fig:PBound} shows the lower bounds in (\ref{eqn:BoundonPe0}) and (\ref{eqn:BoundonPd0}) 
as a function of $\beta J_e$. The lower bounds intersect at $\beta J_c$, i.e., at the criticality of the 
2D homogeneous Ising model in the absence of an external field.

\begin{remark} \label{rem:Uncert}
From (\ref{eqn:BoundonPe0}) and (\ref{eqn:BoundonPd0}) we conclude that
in an arbitrary ferromagnetic Ising model in a nonnegative external field
\begin{equation}
\label{eqn:Uncertainty}
\pi_{\text{p},e}(0)\pi_{\text{d},e}(0) \ge \frac{1}{2},
\end{equation}
which is in the form of an uncertainty principle.
\end{remark}


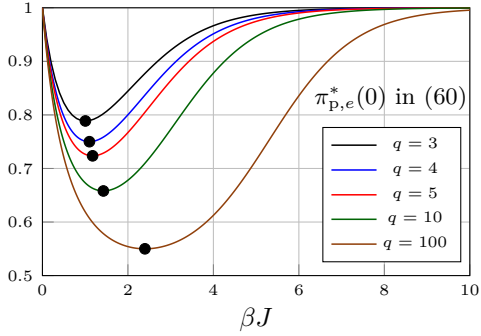
\begin{figure}
\centering
\begin{tikzpicture}
\begin{axis}[
			legend style={at = {(0.98,0.555)} ,font=\tiny},		
			height = 34.0ex,
			width = 48.0ex,
			grid = major,
			tick pos=left, 
			xlabel shift = -2 pt,
			xminorticks = false,	
		    yminorticks = false,	
		    y tick label style={
        /pgf/number format/.cd,
            fixed,
        /tikz/.cd
    		}, 				
			ytick={0.5, 0.6, 0.7, 0.8, 0.9, 1.0},
			xtick={0, 2, 4, 6, 8, 10},
		xlabel= $\beta J$ ={font=\normalsize},
			xmin = 0.0,
			xmax = 10.0,
			ymin = 0.5,
			ymax = 1.0,
			yticklabel style = {font=\tiny,yshift=0.0ex},
            xticklabel style = {font=\tiny,xshift=0.0ex}			
			]

\pgfplotstableread{./PPotts3zero.txt}\mydataone
\pgfplotstableread{./PPotts4zero.txt}\mydatatwo
\pgfplotstableread{./PPotts5zero.txt}\mydatathree
\pgfplotstableread{./PPotts10zero.txt}\mydataeight
\pgfplotstableread{./PPotts100zero.txt}\mydatathirteen
		
		\addplot [
		line width = 0.58 pt,
		 color = black
		]		
		 table[y = Z] from \mydataone;
		 
		 \addplot [
		 line width = 0.58 pt,
		 color = blue
		 ]
		  table[y = Z] from \mydatatwo;	 
		  
		  \addplot [
		 line width = 0.58 pt,
		 color = red
		 ]
		  table[y = Z] from \mydatathree;	 

%
%
%
		
			  	  \addplot [
		 line width = 0.58 pt,
		 color = chocolate1
		 ]
		  table[y = Z] from \mydataeight;	 		  
%
%
%
%
		  
		  	  	  \addplot [
		 line width = 0.58 pt,
		 color = chocolate2
		 ]
		  table[y = Z] from \mydatathirteen;	 		  	  
		    
		\addplot[mark=*] coordinates {(1.005,0.7887)};
		\addplot[mark=*] coordinates {(1.099,0.75)};
		\addplot[mark=*] coordinates {(1.174,0.7236)};
		\addplot[mark=*] coordinates {(1.426,0.658)};
		\addplot[mark=*] coordinates {(2.398,0.55)};
		
		\node at (8.15, 0.83)   {\small{$\pi^{*}_{\text{p}, e}(0)$ in (\ref{eqn:fixedPottsFerzero})}};
%



 \legend{$q=3$, $q=4$, $q=5$, $q=10$, $q=100$};


\end{axis}
\end{tikzpicture}
\vspace{-2.0ex}
\caption{\label{fig:L2}%
The fixed points~(\ref{eqn:fixedPottsFerzero}) as a function of 
$\beta J$ for different values of $q$. 
The filled circles show the fixed points at criticality of the 2D Potts model
located at $\beta J_c = \ln(1+\sqrt{q})$.}
\end{figure}

\section{GENERALIZATION TO NON-BINARY MODELS}
\label{sec:GenP}

We briefly discuss extensions of our mapping to non-binary models, in particular to the $q$-state 
Potts model~\cite{wu1982potts, Baxter:07}. Accordingly, we let $\calA = \ZZ/q\ZZ$ for 
some integer $q \ge 2$. (The binary Ising model is recovered as the special case $q=2$.)

In the absence of an external field, the Hamiltonian of the model is given by
\begin{equation} \label{eqn:HamiltonianIsingPotts}
\HH(\x) = -\sum_{e \in \EE} J_{e}\cdot\delta(y_e).
\end{equation}

From (\ref{eqn:PBoltz}) and (\ref{eqn:HamiltonianIsingPotts}), 
we obtain that in the primal NFG
\begin{equation} 
\label{eqn:PottsJmod}
\psi_{e}(y_e) = \left\{ \begin{array}{ll}
     e^{\beta J_e}, & \text{if $y_e = 0$} \\
     1, & \text{otherwise,}
  \end{array} \right.
\end{equation}
and in the dual NFG, factors are equal to the 1D DFT of (\ref{eqn:PottsJmod}) given by
\begin{equation} 
\label{eqn:PottsDJ}
\tilde \psi_{e}(\tilde y_e) = \left\{ \begin{array}{ll}
      e^{\beta J_e} -1 + q, & \text{if $\tilde y_e = 0$} \\
      e^{\beta J_e} -1, & \text{otherwise,}
  \end{array} \right.
\end{equation}
which is nonnegative if the model is ferromagnetic (i.e., $J_e \ge 0$). See~\cite{AY:2014, MoGo:2018} 
for more details on constructing the primal and the dual NFG of the Potts model, with or without an
external field.

A straightforward generalization Proposition~(\ref{prop:mappingDFT}) gives the mapping 
between $\{\pi_{\text{p}, e}(a)/\psi_{e}(a), a \in \calA\}$
and $\{\pi_{\text{d}, e}(a')/\tilde \psi_{e}(a'), a' \in \calA\}$ via $W_q = \{w_{k,\ell}, k, \ell \in \calA\}$ with 
$w_{k, \ell} = e^{\frac{-2\pi \mathrm{i}}{|\calA|}k\ell}$, where $W_q$ is the $q$-point DFT matrix (i.e., the Vandermonde matrix for 
the roots of unity) and $\mathrm{i}=\sqrt{-1}$ (see~\cite{Brace:99}).

However, due to symmetry in the factors of the primal~(\ref{eqn:PottsJmod}) and the dual Potts 
model~(\ref{eqn:PottsDJ}), we have
\begin{equation} 
\label{eqn:PottsSymPrimal}
\pi_{\text{p}, e}(1)/\psi_e(1) = \ldots = \pi_{\text{p}, e}(q-1)/\psi_e(q-1)
\end{equation}
and
\begin{equation} 
\label{eqn:PottsSymDual}
\pi_{\text{d}, e}(1)/\tilde \psi_e(1) = \ldots = \pi_{\text{d}, e}(q-1)/\tilde \psi_e(q-1).
\end{equation}

Thus, e.g., for $q=3$, the mapping yields
\begin{align}
\label{eqn:PottsMapping}
\frac{\pi_{\text{p}, e}(1)}{\psi_e(1)} & = \frac{\pi_{\text{d}, e}(0)}{\tilde \psi_e(0)} 
+ \frac{\pi_{\text{d}, e}(1)}{\tilde \psi_e(1)}e^{\frac{-2\pi \mathrm{i}}{3}} 
+\frac{\pi_{\text{d}, e}(2)}{\tilde \psi_e(2)}e^{\frac{-4\pi \mathrm{i}}{3}} \notag \\
& = \frac{\pi_{\text{d}, e}(0)}{\tilde \psi_e(0)} - \frac{\pi_{\text{d}, e}(1)}{\tilde \psi_e(1)},
\end{align}
which is real-valued. 


\begin{figure}[t]
\centering
\begin{tikzpicture}
\begin{axis}[
			legend style={at = {(0.98,0.78)} ,font=\tiny},		
			height = 34.0ex,
			width = 48.0ex,
			grid = major,
			tick pos=left, 
			xlabel shift = -2 pt,
			xminorticks = false,	
		    yminorticks = false,	
		    y tick label style={
        /pgf/number format/.cd,
            fixed,
        /tikz/.cd
    		}, 				
			ytick={0, 0.02, 0.04, 0.06, 0.08, 0.1, 0.12},
		xlabel= $\beta J$ ={font=\normalsize},
			xmin = 0.0,
			xmax = 10.0,
			ymin = 0.0,
			ymax = 0.12,
			yticklabel style = {font=\tiny,yshift=0.0ex},
            xticklabel style = {font=\tiny,xshift=0.0ex}			
			]

\pgfplotstableread{./PPotts3one.txt}\mydataone
\pgfplotstableread{./PPotts4one.txt}\mydatatwo
\pgfplotstableread{./PPotts5one.txt}\mydatathree
\pgfplotstableread{./PPotts10one.txt}\mydataeight
\pgfplotstableread{./PPotts100one.txt}\mydatathirteen
		
		\addplot [
		line width = 0.58 pt,
		 color = black
		]		
		 table[y = Z] from \mydataone;
		 
		 \addplot [
		 line width = 0.58 pt,
		 color = blue
		 ]
		  table[y = Z] from \mydatatwo;	 
		  
		  \addplot [
		 line width = 0.58 pt,
		 color = red
		 ]
		  table[y = Z] from \mydatathree;	 

%
%
%
		
			  	  \addplot [
		 line width = 0.58 pt,
		 color = chocolate1
		 ]
		  table[y = Z] from \mydataeight;	 		  
%
%
%
%
		  
		  	  	  \addplot [
		 line width = 0.58 pt,
		 color = chocolate2
		 ]
		  table[y = Z] from \mydatathirteen;

		\addplot[mark=*] coordinates {(1.005,0.10565)};
		\addplot[mark=*] coordinates {(1.099,0.08333)};
		\addplot[mark=*] coordinates {(1.174,0.0691)};
		\addplot[mark=*] coordinates {(1.426,0.038)};
		\addplot[mark=*] coordinates {(2.398,0.004545)};

		\node at (8.15, 0.105)   {\small{$\pi^{*}_{\text{p}, e}(1)$ in (\ref{eqn:fixedPottsFerone})}};

  \legend{$q=3$, $q=4$, $q=5$, $q=10$, $q=100$};


\end{axis}
\end{tikzpicture}
\vspace{-2.0ex}
\caption{\label{fig:L3}%
Everything as in Fig.~\ref{fig:L2} but for $\pi^{*}_{\text{p}, e}(1)$ in~(\ref{eqn:fixedPottsFerone}).}
\end{figure}
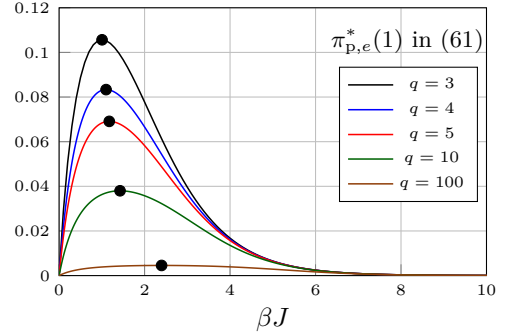

For a homogeneous and ferromagnetic $q$-state Potts model, the fixed points are
given by
\begin{equation}
\label{eqn:fixedPottsFerzero}
\pi^{*}_{\text{p},e}(0) = \frac{e^{\beta J}(e^{\beta J} -1 + q)}{e^{2\beta J}- 2(1-q)e^{\beta J} + 1 -q}
\end{equation}
and
\begin{equation}
\label{eqn:fixedPottsFerone}
\pi^{*}_{\text{p},e}(t) = \frac{e^{\beta J}-1}{e^{2\beta J}-2(1-q)e^{\beta J} + 1 - q}
\end{equation}
for $t \in \{1, 2, \ldots, q-1\}$.

Figs.~\ref{fig:L2} and \ref{fig:L3} show the fixed point~(\ref{eqn:fixedPottsFerzero}) 
and (\ref{eqn:fixedPottsFerone}) as a function of $\beta J$. 
Like the Ising model, the minimum of $\pi^{*}_{\text{p},e}(0)$ 
is attained at the criticality of the 
2D Potts model 
located at $\beta J_c = \ln(1+\sqrt{q})$~\cite{wu1982potts}. 

It is easy to show that at criticality and in the many-component limit (i.e., as $q \to \infty$), we have
\begin{equation}
\lim_{q \to \infty} \pi^{*}_{\text{p},e}(0) = \frac{1}{2}
\end{equation}

\begin{remark} \label{rem:comp}
Transforming marginals from one domain to the other requires a matrix-vector multiplication with 
computational complexity $\mathcal{O}(|\calA|^2)$. 
However, when there is symmetry in the factors, as in~(\ref{eqn:IsingJmod}) and~(\ref{eqn:PottsJmod}), the complexity can be 
reduced to $\mathcal{O}(|\calA|)$.
\end{remark}

\begin{remark} \label{rem:hint}
In binary models, factors in the dual NFG can in general take negative values, 
and in nonbinary models, the factors can be complex-valued. 
In such cases a 
valid PMF can no longer be defined in the 
dual domain. The mappings remain nevertheless valid; but for 
\emph{marginal functions} (instead of marginal densities) 
of a global function with a factorization given by~(\ref{eqn:Pd}). 
\end{remark}

\section{NUMERICAL EXPERIMENTS}

In both domains
estimates of marginal densities can be obtained via Monte Carlo methods or via 
variational 
algorithms~\cite{christian1999monte, murphy:2012}.
We only consider the subgraphs-world process (SWP) and two variational algorithms, the 
belief propagation (BP) and 
the tree expectation propagation (TEP), for the Ising model. 
Estimated marginals in the dual domain are then transformed 
all together to the primal domain via (\ref{eqn:MapG}) and(\ref{eqn:PottsMapping}). 
In all experiments,  the exact marginal 
densities are computed via the junction tree algorithm implemented in \cite{Mooij:2010}.

The choice of methods and the models is far from exhaustive -- our goal is 
to show the advantage of using the mappings in approximating marginal densities in similar settings.

In our first experiment, we consider a 2D homogeneous Ising model, in a constant external field $\beta H = 0.15$, with 
periodic boundaries, and with size $N = 6\times6$. For this model, BP and TEP 
in the primal and in the dual domains give virtually indistinguishable approximations. We also apply 
SWP using $10^5$ samples.
\Fig{fig:IsingHom} shows the relative error in estimating $\pi_{\text{p}, e}(0)$ as 
a function of $\beta J$, 
where SWP (which operates in the dual NFG) gives good estimates in the whole range. 

Compared to variational algorithms, convergence of the SWP is slow; moreover,
SWP is only applicable to ferromagnetic Ising models in a positive field. (Indeed, in order to have 
an irreducible Markov chain in the SWP the external field needs to
be non-zero~\cite[Chapter 8]{Welsh:93}). In the rest of the experiments, we consider Ising models in the absence of an
 external field, and only compare the efficiency of variational algorithms employed in 
 the primal and in the dual domains.

\begin{figure}[h!!!]
\centering
\begin{tikzpicture}
\begin{semilogyaxis}[
			legend style={at = {(0.52,0.36)} ,font=\tiny},		
			height = 37.0ex,
			width = 49.0ex,
			grid = major,
			tick pos=left, 
			xlabel shift = -2 pt,
			xminorticks = false,	
		    yminorticks = false,
    		x tick label style={
        /pgf/number format/.cd,
            fixed,
            fixed zerofill,
            precision=2,
        /tikz/.cd
    		}, 						
			ytick={1e-5, 1e-4, 1e-3, 1e-2, 1e-1, 1e0},
			yticklabels = {$10^{-5}$, $10^{-4}$, $10^{-3}$, $10^{-2}$, $10^{-1}$, $1$},
			xtick={0.05, 0.15, 0.25, 0.35, 0.45, 0.55, 0.65},
			xlabel= $\beta J_x$ ={font=\normalsize},
			xmin = 0.05,
			xmax = 0.65,
			ymin = 1e-5,
			ymax = 1e0,
			yticklabel style = {font=\tiny,yshift=0.0ex},
            xticklabel style = {font=\tiny,xshift=0.0ex}	]

\addplot [mark size=1.3, blue, mark=triangle*, densely dashed, line width = 0.65 pt] table[x={J}, y={BP}] {./RelativeErrorPFully.txt};
\addplot [mark size=1.2, red, mark=square*, densely dotted, line width = 0.65 pt] table[x={J}, y={TP}] {./RelativeErrorPFully.txt};
\addplot [mark size=1.2, black, mark=*, densely dashdotted, line width = 0.65 pt] table[x={J}, y={BP}] {./RelativeErrorDFully.txt};

\legend{BP primal, TEP primal, BP/TreeEP dual};

\end{semilogyaxis}
\end{tikzpicture}
\vspace{-2.0ex}
\caption{\label{fig:IsingFully}%
Average relative error in estimating $\pi_{\text{p}, e}(0)$ in a fully-connected 
Ising model with $N = 10$. Coupling parameters are chosen uniformly and independently between 0.05 and $\beta J_x$ 
denoted by the $x$-axis.}
\end{figure}
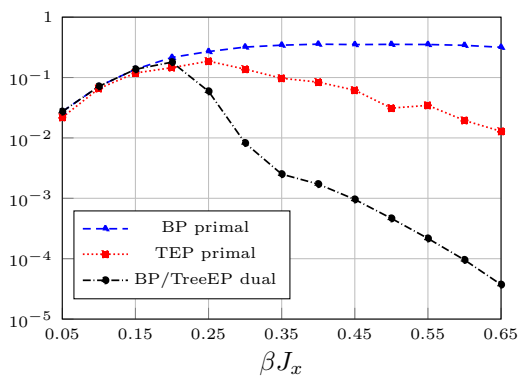

In the second experiment, we consider a 2D Ising model with size $N = 6\times6$ and with 
periodic boundaries. Couplings are chosen randomly according
to a half-normal distribution, i.e., $\beta J_e = |\beta J'_e|$ with $\beta J'_e \overset{\text{i.i.d.}}{\sim} \calN(0, \sigma^2)$. 
\Fig{fig:IsingGauss} shows the average relative error in estimating the marginal 
density $\pi_{\text{p}, e}(0)$ as a function of $\sigma^2$, where the results are averaged over 200 independent realizations.
We consider a fully-connected Ising model with $N = 10$ in our last experiment. Couplings are 
chosen randomly 
according to $\beta J_e \overset{\text{i.i.d.}}{\sim} \calU [0.05, \beta J_x]$, i.e., uniformly between 0.05 
and $\beta J_x$
denoted by the $x$-axis. The average relative error over 50 independent realizations 
is illustrated in~\Fig{fig:IsingFully}.

\begin{figure}[t]
\centering
\begin{tikzpicture}
\begin{semilogyaxis}[
			legend style={at = {(0.53,0.385)} ,font=\tiny},		
			height = 37.0ex,
			width = 49.0ex,
			grid = major,
			tick pos=left, 
			xlabel shift = -2 pt,
			xminorticks = false,	
		    yminorticks = false,
    		x tick label style={
        /pgf/number format/.cd,
            fixed,
            fixed zerofill,
            precision=2,
        /tikz/.cd
    		}, 						
			ytick={1e-6, 1e-5, 1e-4, 1e-3, 1e-2, 1e-1},
			yticklabels = {$10^{-6}$, $10^{-5}$, $10^{-4}$, $10^{-3}$, $10^{-2}$},
			xtick={0.05, 0.15, 0.25, 0.35, 0.45, 0.55, 0.65, 0.75, 0.85},
			xlabel= $\beta J$ ={font=\normalsize},
			xmin = 0.05,
			xmax = 0.75,
			ymin = 1e-6,
			ymax = 1e-2,
			yticklabel style = {font=\tiny,yshift=0.0ex},
            xticklabel style = {font=\tiny,xshift=0.0ex}	]

\addplot [mark size=1.3, blue, mark=triangle*, densely dashed, line width = 0.65 pt] table[x={J}, y={BP}] {./IsingREH0.15.txt};
\addplot [mark size=1.2, red, mark=square*, densely dotted, line width = 0.65 pt] table[x={J}, y={TP}] {./IsingDREH0.15.txt};
\addplot [mark size=1.3, chocolate1, mark= star, dashdotted, line width = 0.65 pt] table[x={J}, y={RE}] {./GibbsDual2.txt};

\legend{BP primal/dual, TEP primal/dual, SWP};

\end{semilogyaxis}
\end{tikzpicture}
\vspace{-2.0ex}
\caption{\label{fig:IsingHom}%
Relative error as a function of  $\beta J$ in estimating $\pi_{\text{p}, e}(0)$ of a homogeneous Ising model 
in a constant external field $\beta H = 0.15$, with periodic boundaries, and with size $N = 6\times6$.}
\end{figure}
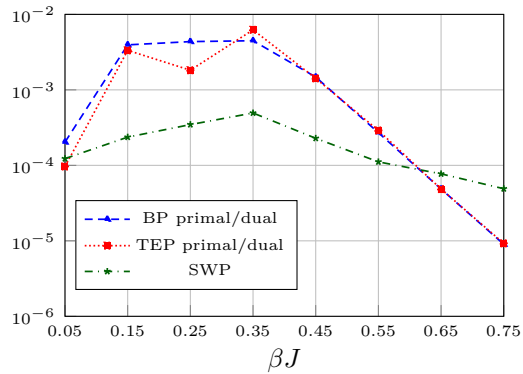
\begin{figure}
\centering
\begin{tikzpicture}
\begin{semilogyaxis}[
			legend style={at = {(0.52,0.355)} ,font=\tiny},		
			height = 37.0ex,
			width = 49.0ex,
			grid = major,
			tick pos=left, 
			xlabel shift = -2 pt,
			xminorticks = false,	
		    yminorticks = false,
    		x tick label style={
        /pgf/number format/.cd,
            fixed,
            fixed zerofill,
            precision=2,
        /tikz/.cd
    		}, 						
			ytick={1e-5, 1e-4, 1e-3, 1e-2, 1e-1, 1e0},
			yticklabels = {$10^{-5}$, $10^{-4}$, $10^{-3}$, $10^{-2}$, $10^{-1}$, $1$},
			xtick={0.05, 0.25, 0.45, 0.65, 0.85, 1.05, 1.25, 1.45, 1.65, 1.85},
			xlabel= $\sigma^2$ ={font=\normalsize},
			xmin = 0.05,
			xmax = 1.85,
			ymin = 1e-5,
			ymax = 1e0,
			yticklabel style = {font=\tiny,yshift=0.0ex},
            xticklabel style = {font=\tiny,xshift=0.0ex}	]

\addplot [mark size=1.3, blue, mark=triangle*, densely dashed, line width = 0.65 pt] table[x={J}, y={BP}] {./RelativeErrorPIsingGauss.txt};
\addplot [mark size=1.2, red, mark=square*, densely dotted, line width = 0.65 pt] table[x={J}, y={TP}] {./RelativeErrorPIsingGauss.txt};
\addplot [mark size=1.2, black, mark=*, densely dashdotted, line width = 0.65 pt] table[x={J}, y={BP}] {./RelativeErrorDIsingGauss.txt};

\legend{BP primal, TEP primal, BP/TreeEP dual};

\end{semilogyaxis}
\end{tikzpicture}
\vspace{-2.0ex}
\caption{\label{fig:IsingGauss}%
Average relative error in estimating $\pi_{\text{p}, e}(0)$ of an Ising model with 
periodic boundaries and with size $N = 6\times6$. Couplings are chosen randomly according
to a half-normal distribution with variance $\sigma^2$.}
\end{figure}
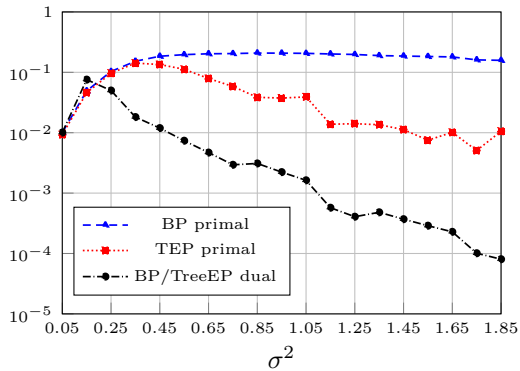


In both experiments, BP and TEP provide
close approximations in the dual domain, therefore only 
BP results are reported.
Figs.~\ref{fig:IsingGauss} and~\ref{fig:IsingFully} show that for 
$\sigma^2 > 0.25$ and $\beta J_x > 0.20$, BP in the dual NFG can significantly improve the quality of 
estimates -- even by more than two orders of magnitude in terms of relative error. 

\section{CONCLUSION}

We proved that marginals densities of a
primal NFG and the corresponding marginal densities of its dual NFG are related  
via local mappings. The mapping provides a simple procedure to transform 
simultaneously the estimated marginals 
from one domain to the other. Furthermore, the mapping relies on no assumptions
on the size or on the topology of the graphical model. 
Our numerical experiments show that estimating the marginals in the dual NFG 
can sometimes significantly improve the quality of
approximations 
in terms of relative error. 
In the special case of the ferromagnetic Ising model in a positive external field, there is indeed a rapidly mixing Markov
chain (the subgraphs-world process) to generate configurations in the dual domain. 

\subsubsection*{Acknowledgements}

The author is extremely grateful to G.~D.~Forney, Jr., for his comments and for his continued support. 
The author also wishes to thank J. Dauwels, P. Vontobel, V.~G\'{o}mez, and B.~Ghojogh for their comments.



%
%

\bibliography{mybib}
\end{document}